\UseRawInputEncoding
\documentclass[a4paper,10pt,oneside]{article}
\usepackage[utf8]{inputenc}
\usepackage{twomod,amsmath,epsfig,times,url,hyperref}
\usepackage[T1]{fontenc}
\usepackage{booktabs}       
\usepackage{caption}        
\usepackage{cuted}
\usepackage{siunitx}        

\title{How Lightweight Can a Vision Transformer Be}

\author{
 Jen Hong Tan \\
  Data Science and Artificial Intelligence Lab (DSAIL)\\
  Health Services Research Unit\\
  Singapore General Hospital \\
}

\begin{document}
\ninept
\maketitle

\begin{sloppy}
\begin{abstract}
In this paper, we explore a strategy that uses Mixture-of-Experts (MoE) to streamline, rather than augment, vision transformers. Each expert in an MoE layer is a SwiGLU feedforward network, where $\mathbf{V}$ and $\mathbf{W_2}$ are \textbf{\textit{shared}} across the layer. No complex attention or convolutional mechanisms are employed. Depth-wise scaling is applied to progressively reduce the size of the hidden layer and the number of experts is increased in stages. Grouped query attention is used. We studied the proposed approach with and without pre-training on small datasets and investigated whether transfer learning works at this scale. We found that the architecture is competitive even at a size of 0.67M parameters.
\end{abstract}

\section{Introduction}
\label{sec:intro}

In real-world applications of computer vision, such as edge intelligence, small and performant models are still preferred to overcome computational challenges \cite{LIU2022297}. Vision Transformers (ViTs) \cite{dosovitskiy2020image} have achieved remarkable results, but their performance drops significantly when the model size and dataset are small \cite{Zheng2023}. Consequently, there are studies investigating lightweight vision transformers that perform well on mid-size datasets. Almost all of these studies either use new types of attention blocks \cite{pan2022edgevits, pvt2021wang} or integrate convolutional mechanisms \cite{mehta2022mobilevit} into their architectures.

On the other hand, Tan \cite{tan2024pretraining} has shown that by employing Masked Auto-Encoder (MAE) \cite{he2022masked} as a pre-training strategy, it is possible to get ViT to learn effectively from small datasets. In that work, the ViT consists of 12 transformer encoder layers, each containing a multi-head attention component and a feedforward network. The feedforward network consists of two linear layers: the first expands the output to \textbf{twice}, rather than four times, the embedding size, and the second reduces the output back to the embedding size. To further lighten the model, reducing the expanded output size in the middle of the feedforward network can help, but excessive reduction can negatively affect model performance.

With these considerations in mind, we designed an architecture that uses Mixture-of-Experts (MoE) \cite{shazeer2017} to streamline vision transformers. In our architecture, each expert in a MoE layer is formed by a SwiGLU \cite{swiglu} feedforward network. By design, SwiGLU is heavier in terms of parameters compared to a typical multi-layer perceptron. However, with several experts in a MoE layer, we are able to make the hidden size in SwiGLU smaller than the embedding size without negatively affecting model performance. Furthermore, we share 2 out of the 3 linear transformations in each SwiGLU across the layer. This helps to significantly lower the parameter count while maintaining the strength of MoE. Beyond that, to further reduce the number of parameters, we progressively increase the number of experts in the MoE in stages, while linearly reducing the hidden size by depth, borrowing the idea from depth-wise scaling \cite{mehta2021delight}. Lastly, we use grouped query attention \cite{ainslie2023gqa} to keep the parameter count low. Source code will be provided in near future.

\section{Method}
\label{sec:method}

Our proposed approach consists of two parts: \textbf{mLiT} and \textbf{mmLiT}. \textbf{mLiT} is a \textbf{m}ixture-of-experts based \textbf{Li}ghtweight vision \textbf{T}ransformer, and \textbf{mmLiT} is a \textbf{mLiT} pre-trained and fine-tuned using \textbf{m}asked auto-encoder pretraining strategy.

\subsection{mLiT}
\label{ssec:mlit}

Similar to the original ViT \cite{dosovitskiy2020image}, we start by dividing each image into non-overlapping patches. Each of these patches is linearly transformed into a set of embeddings, augmented by learnable positional embeddings. The processed embeddings go through a series of MoE-based transformer encoder layers. Figure \ref{fig:mLiT} shows the overall structure of our transformer encoder layer .

\begin{figure} 
    \centering
    \includegraphics[trim=10cm 4cm 10cm 3cm, clip, scale=0.5]{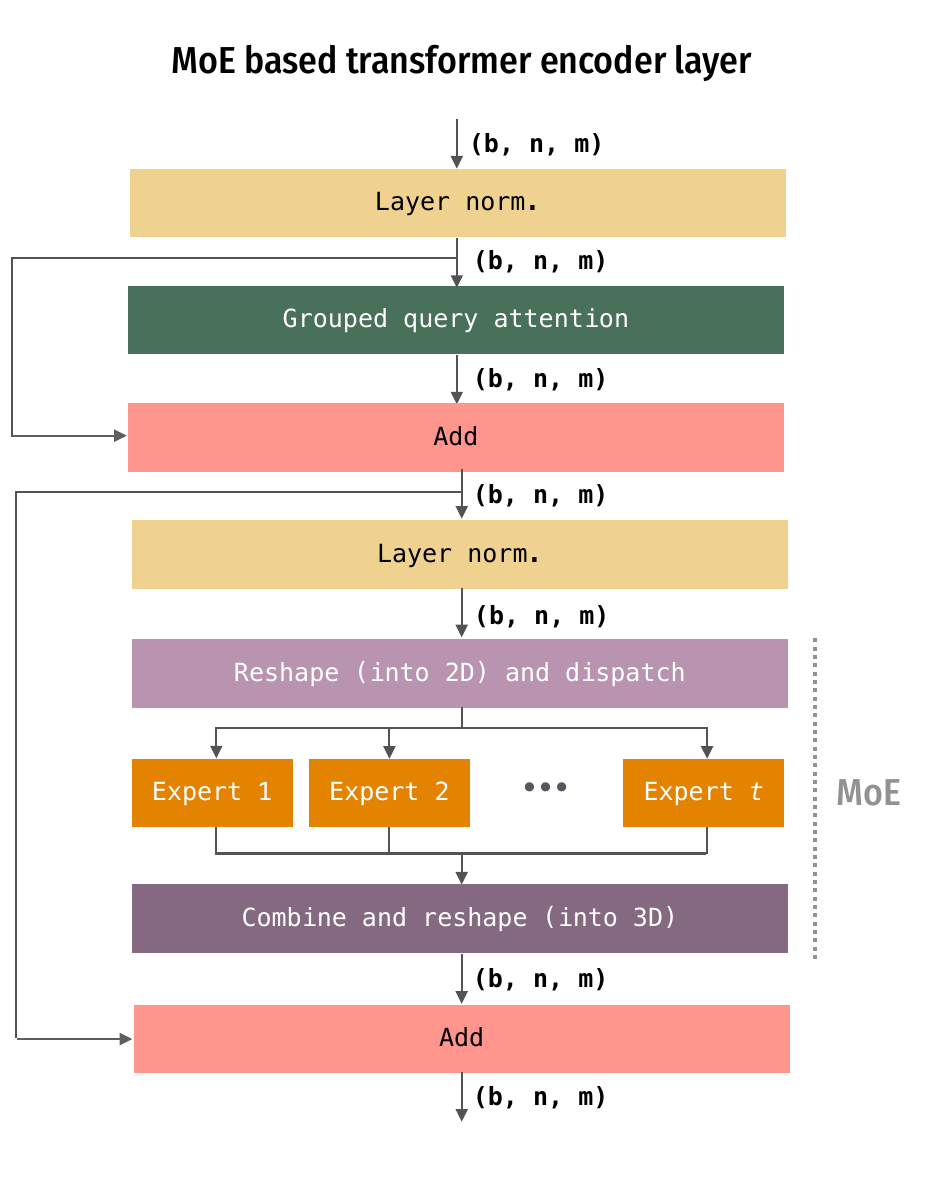}
    \caption{The structure of an MoE based transformer encoder layer. Each expert is a SwiGLU feedforward network. $b$ stands for batch size, $n$ for the number of embeddings and $m$ for embedding size. In this layer there are $t$ number of experts}
    \label{fig:mLiT}
\end{figure}

\begin{figure*} 
    \centering
    \includegraphics[trim=0cm 9cm 0cm 6.5cm, clip, scale=0.45]{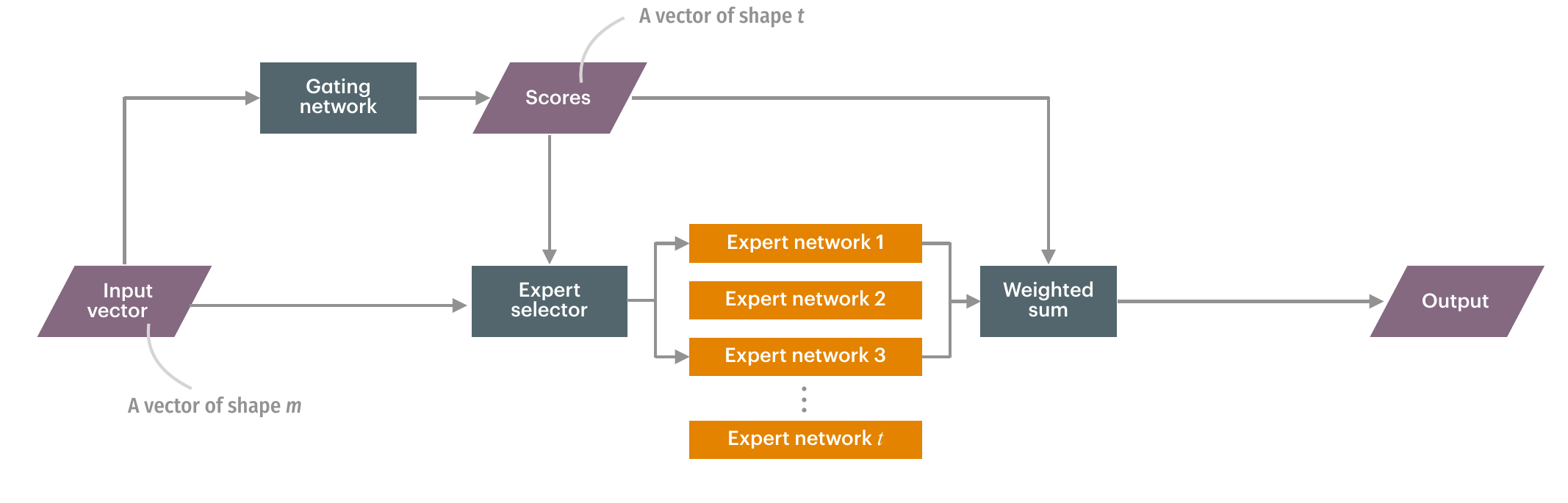}
    \caption{The working of MoE assuming the input is a vector. In this example, expert 1 and expert 3 receive the input as directed by the gating network}
    \label{fig:moe_vector}
\end{figure*}

\subsubsection{Grouped Query Attention (GQA)}

Grouped query attention divides query heads into $G$ groups, with each group sharing a single key head and value head. For instance, GQA-1, with a single group, is equivalent to multi-query attention (MQA); while GQA-$H$, with groups equal to the number of heads ($H$), is equivalent to multi-head attention (MHA). An intermediate number of groups results in a model that is higher quality than MQA but faster than MHA \cite{ainslie2023gqa}, providing a favorable trade-off between performance and memory bandwidth.

\subsubsection{Mixture of Experts (MoE)}

A single Mixture-of-Experts layer comprises $t$ number of expert networks $E_1, E_2, \cdots, E_t$, with a gating network $G$ that directs each input vector/embedding to the relevant expert(s) in the layer. For each vector or embedding, the output of $G$ is a sparse $t$-dimensional vector. Figure \ref{fig:moe_vector} shows an overview of the working of MoE when the input is a vector.

Let's denote a vector/embedding by $x$, the output of the $i$-th expert by $E_i(x)$, and the $i$-th component of the output of the gating network $G(x)$ by $G(x)_i$. The output $y$ of the MoE is given by:

\begin{equation}
y = \sum_{i=1}^t G(x)_i E_i(x)
\vspace{0.8em}
\end{equation}

As the output $G(x)$ is sparse, in MoE, no computation is performed on the $j$-th expert when $G(x)_j = 0$. In many applications, however, the input to the layer is generally of shape $(b,n,m)$, where $b$, $n$, and $m$ denote batch size, the number of embeddings and the embedding size, respectively. Therefore, it is common to reshape the input tensor into 2D \cite{rau2019moe} before the MoE layer. Figure \ref{fig:moe_tensor} shows the workflow of the actual implementation of MoE.

\begin{figure*} 
    \centering
    \includegraphics[trim=0cm 9cm 0cm 6cm, clip, scale=0.45]{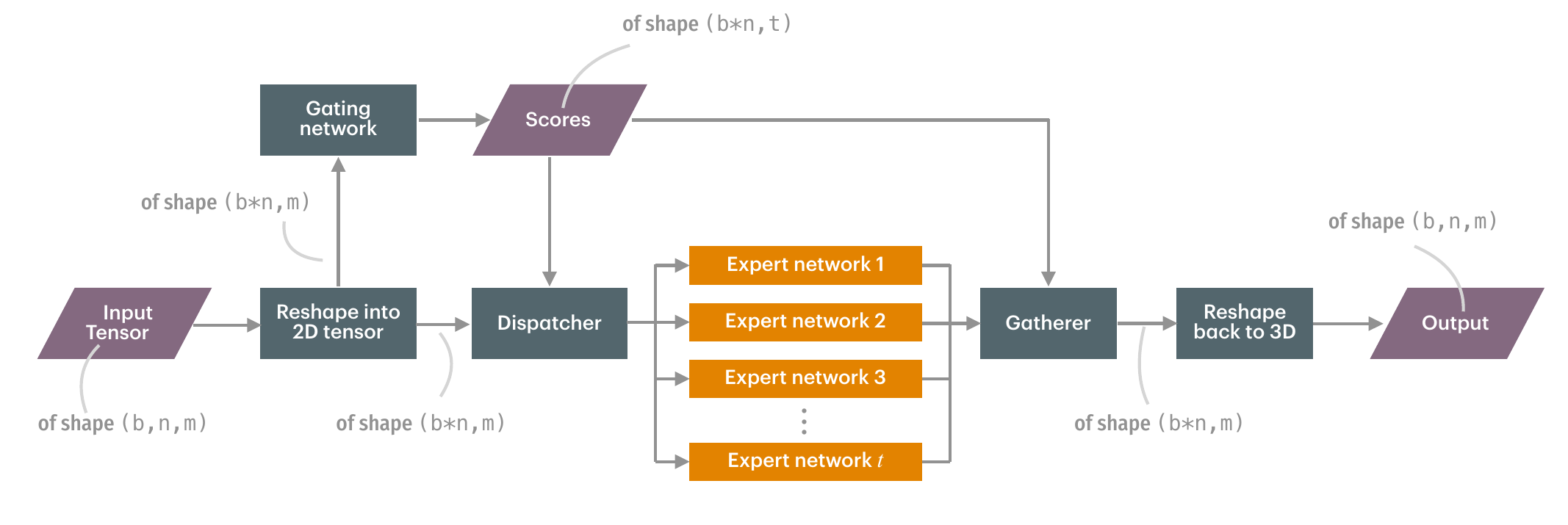}
    \caption{The actual implementation of MoE. Unlike the case when the input is just a vector, when $b \cdot n$ embeddings go into the layer, almost all the expert networks will receive some embeddings sent by the dispatcher}
    \label{fig:moe_tensor}
\end{figure*}

\textbf{Noisy Top-K Gating Network.} We follow the proposal by Shazeer et. al. \cite{shazeer2017} that adds sparsity and noise to a softmax gating mechanism. Assume a reshaped input tensor $\mathbf{x}$ with a shape $(b \cdot n, m)$ and two trainable weight matrices $\mathbf{W}_g$ and $\mathbf{W}_{noise}$ , both of shape $(m, t)$, where $t$ is the total number of experts in the layer and $k$ is the number of expert(s) to be selected. The output of the gating network is given by

\begin{equation}
G(\mathbf{x}) = \mathrm{Softmax}_k \left( H(\mathbf{x}) \right)
\end{equation} where \begin{equation}
H(\mathbf{x}) = \mathbf{x}\mathbf{W}_g + \mathrm{randn\_like({\mathbf{x}\mathbf{W}_g}}) \odot \mathrm{Softplus}\left(  \mathbf{x}\mathbf{W}_{noise} \right)
\vspace{0.8em}
\end{equation}

$\odot$ denotes the element-wise product. $\mathrm{randn\_like}$ generates a tensor filled with random numbers from a normal distribution with mean 0 and variance 1, and the shape of the tensor is equal to the shape of the output from $\mathbf{x} \mathbf{W}_g$. $\mathrm{Softmax}_k$ applies softmax (row-wise) only on the top $k$ elements, with the rest set to 0 in the output. The equation below illustrates how $\mathrm{Softmax}_k$ works when $k$ is 2 and $t$ (the number of experts) is 4:

\begin{strip}
\begin{equation}
\mathrm{Softmax}_k \left( \left[
\begin{array}{rrrr}
4.4742 & -5.6365 & 6.8226 & 0.9960 \\
3.5298 & 2.3049 & 1.2113 & -1.3946 \\
-2.2414 & 0.3925 & 1.6676 & -1.9253 \\
\end{array}
\right]\right) =
\begin{bmatrix}
0.0872 & 0.0000 & 0.9128 & 0.0000 \\
0.7729 & 0.2271 & 0.0000 & 0.0000 \\
0.0000 & 0.2184 & 0.7816 & 0.0000 \\
\end{bmatrix}
\vspace{0.8em}
\end{equation}
\end{strip}

\textbf{Losses for MoE.} To encourage all experts to have equal importance, two loss functions are introduced: load balancing loss and importance loss. To calculate load balancing loss, we first determine the following probability by 

\begin{equation}
P(\mathbf{x}) = \Phi \left( \frac{\mathbf{x}\mathbf{W}_g- \Psi(H(\mathbf{x}))}{\mathrm{Softplus}(\mathbf{x}\mathbf{W}_{noise})} \right)
\vspace{0.8em}
\end{equation} where $\Phi$ is the cumulative distribution function of a standard normal distribution. To calculate the output of $\Psi$, we let $\mathbf{H}$ denote the output of $H(\mathbf{x})$, which has a shape of $(b \cdot n, t)$, with each element in $\mathbf{H}$ denoted by $h_{r,c}$ . Furthermore, we let $\mathbf{H}_r$ denote a row by index $r$ in $\mathbf{H}$. Similarly, $\psi_{r,c}$ denotes each element in the output of $\Psi(H(\mathbf{x}))$. If we let $l^{k}_r$ and $l^{k+1}_r$ denote the $k^{th}$ and $k+1^{th}$ largest values respectively for a row $r$ in $\mathbf{H}$, $l^{k}_r > l^{k+1}_r$,  then

\begin{equation}
\psi_{r,c} = \left\{
  \begin{array}{ll}
    l^{k+1}_r \mbox{ in } \mathbf{H}_r & \mbox{ if } h_{r,c} \geq l^{k}_r \\
    l^{k}_r \mbox{ in } \mathbf{H}_r & \mbox{ if } h_{r,c} < l^{k}_r \\
  \end{array}
\right.
\end{equation} where $k$ is the amount of expert(s) to be selected in the layer. As an example, assume we have 

\begin{equation}
\mathbf{H}= \left[ \begin{array}{rrrr}
4.4742 & -5.6365 & 6.8226 & 0.9960 \\
3.5298 & 2.3049 & 1.2113 & -1.3946 \\
-2.2414 & 0.3925 & 1.6676 & -1.9253 \\
\end{array} \right]
\vspace{0.8em}
\end{equation} and $k$ is 2, then for each row in $\mathbf{H}$ (\textit{with row index starting from 0}), we have

\begin{equation}
\begin{array}{crr}
	\hline
	\mbox{row} & k^{th} \mbox{ largest}      & k+1^{th} \mbox{ largest}     \\
	\hline
	0 & 4.4742 & 0.9960 \\
	1 & 2.3049 & 1.2113 \\
	2 & 0.3925 & -1.9253 \\
	\hline
\end{array}
\vspace{1em}
\end{equation} The output of $\Psi$ is

\begin{equation}
\Psi(\mathbf{H})= \left[ \begin{array}{rrrr}
0.9960 & 4.4742 & 0.9960 & 4.4742 \\
1.2113 & 1.2113 & 2.3049 & 2.3049 \\
0.3925 & -1.9253 & -1.9253 & 0.3925 \\
\end{array} \right]
\vspace{0.8em}
\end{equation} Lastly, with the probability calculated, the load balancing loss is given by: 

\begin{equation}
L_{load}(\mathbf{x})= w_{load} \times \mathrm{CV}\left( \mathrm{sum}_0 (P(\mathbf{x}))\right)
\vspace{0.8em}
\end{equation} where $w_{load}$ is a hand-tuned scaling factor, $\mathrm{CV}$ is the function to calculate coefficient of variation for the input, and $\mathrm{sum}_0$ performs column-wise summation on $P(\mathbf{x})$.To determine importance loss, we simply do

\begin{equation}
L_{importance} = w_{importance} \times \mathrm{sum}_0 (G(\mathbf{x}))
\vspace{0.8em}
\end{equation} and the total loss is 

\begin{equation}
L = L_{importance}+L_{load}
\vspace{0.8em}
\label{eq:moe_loss}
\end{equation} In our implementation, we set $w_{importance} = w_{load} = 1 \times 10^{-2}$.

\textbf{SwiGLU.} Each expert in a MoE layer is a SwiGLU FeedForward Network (FFN). Assume we have a tensor $\mathbf{x}$ of shape $(n, m)$, where $n$ is the number of embeddings and $m$ is the embedding size, we define $\mathbf{W}, \mathbf{V}$ and $\mathbf{W_2}$ of shape $(m,d_h)$, $(m, d_h)$ and $(d_h, m)$ respectively. $d_h$ is the hidden size. The output of the network is given by:

\begin{equation}
\mathrm{FFN_{SwiGLU}}(\mathbf{x}) = \left(\mathrm{silu}\left(\mathbf{xW} \right) \odot \left(\mathbf{xV} \right) \right)\mathbf{W}_2
\vspace{0.8em}
\end{equation} Biases are omitted in the above for $\mathbf{W}, \mathbf{V}$ and $\mathbf{W_2}$. $\mathrm{silu}$ is the swish function:

\begin{equation}
\mathrm{silu}(x) = x * \sigma(x)
\vspace{0.8em}
\end{equation} where $\sigma(x)$ is the logistic sigmoid. Dropout is applied on the output of SwiGLU FFN.

In \textbf{mLiT}, $\mathbf{V}$ and $\mathbf{W_2}$ and their corresponding biases in each expert are shared in a MoE layer.

\subsubsection{Depth-wise Scaling}

Our vision transformer does not have a fixed hidden size $d_h$ across the transformer encoder layers. Instead, the hidden size is linearly reduced from the first layer to the last layer. Let $d^{\mathrm{first}}_h$ and $d^{\mathrm{last}}_h$ denote the hidden size of the first and the last layer respectively, the hidden size of any layer is given by

\begin{equation}
l^i_h = \left\lfloor \frac{(L_E - 1)-i}{L_E-1}\left(d^{\mathrm{first}}_h-d^{\mathrm{last}}_h \right) \right\rfloor + d^{\mathrm{last}}_h
\label{eq:hid_size}
\end{equation} where $L_E$ is the total number of the transformer encoder layers, $\lfloor \cdot \rfloor$ is floor operator, and $i$ is a layer index starting from 0.

Furthermore, the number of experts in \textbf{mLiT} is increased at different stages, specifically from 3 to 5 after every 3, 4, or 5 layers, depending on the model size. See Figure \ref{fig:mLiT-XXS} for more detail.

\begin{figure} 
    \centering
    \includegraphics[trim=0cm 0.5cm 0cm 0.2cm, clip, scale=0.45]{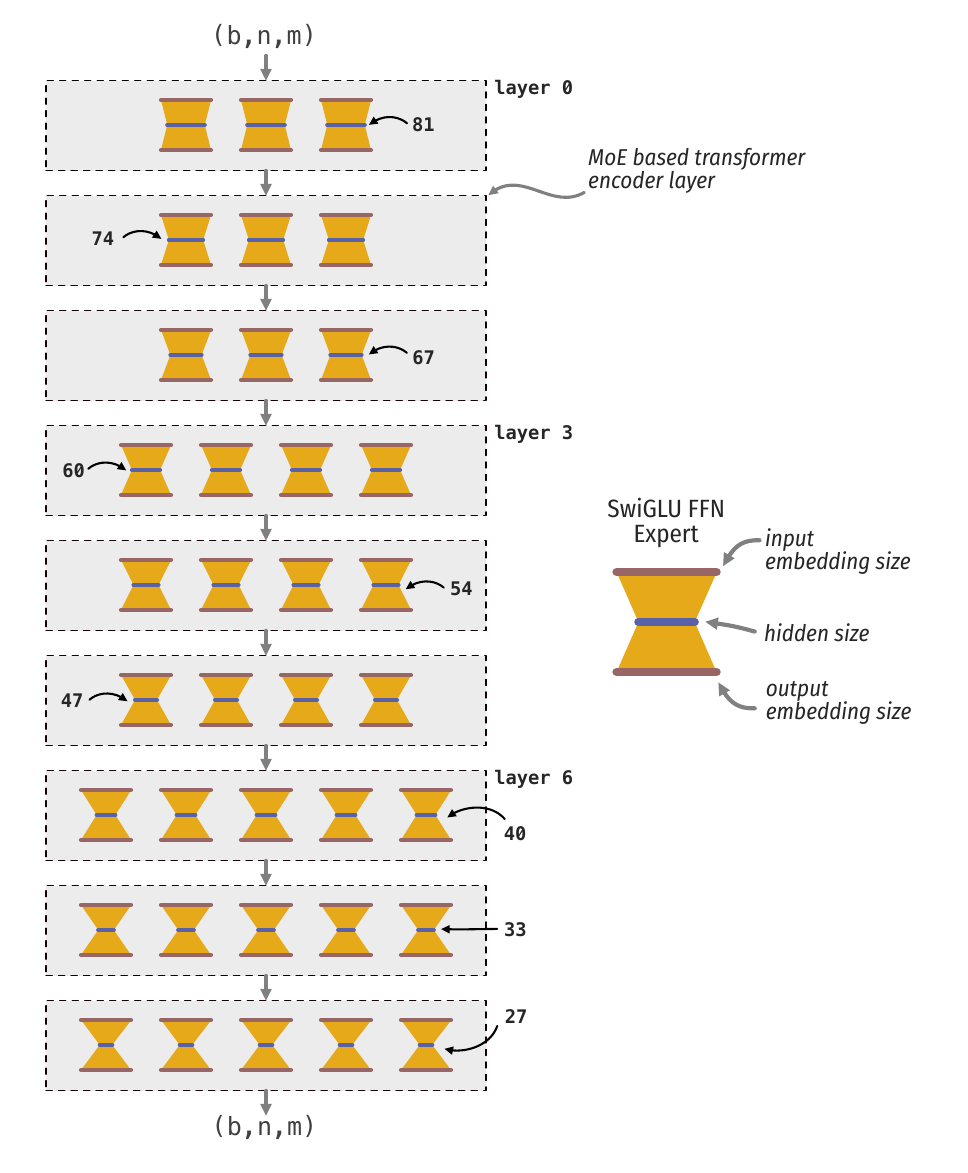}
    \caption{The overall architecture of \textbf{mLiT}. The linear transformation and positional embeddings before the first layer are not included in the illustration. $b$ stands for batch size, $n$ for the number of embeddings and $m$ for embedding size. The input and the output embedding size of each SwiGLU FFN expert are equal to the embedding size $m$. The model in this figure has 9 MoE based transformer encoder layers, with the number of experts increased by 1 at layer 3 and layer 6. $d^{\mathrm{first}}_h$ and $d^{\mathrm{last}}_h$ are 81 and 27, respectively. The hidden size at each layer is calculated using Equation \ref{eq:hid_size}.}
    \label{fig:mLiT-XXS}
\end{figure}

\subsection{mmLiT}

In \textbf{mmLiT}, we apply \textbf{m}asked auto-encoder on \textbf{mLiT}. We follow strictly the process outlined in \cite{tan2024pretraining}, which is based on the original MAE paper \cite{he2022masked}. The masking ratio is set to 0.75. However, unlike in \cite{tan2024pretraining}, we did not incorporate a separable learnable positional embeddings at the decoder's input (See Figure \ref{fig:mmLiT_layers}). Similar to \cite{tan2024pretraining}, we compute the loss of the auto-encoder as a sum of the loss on masked patches and an additional, discounted loss on unmasked patches. The total loss is given by

\begin{equation}
\mathrm{Loss} = \mathrm{MSE}_{\mathrm{masked}} + \alpha \cdot \mathrm{MSE}_{\mathrm{unmasked}} + \beta \cdot L
\label{eq:loss}
\end{equation} where $\alpha$ denotes the discounting factor, $\beta$ is the loss coefficient for MoE. $L$ is calculated from Equation \ref{eq:moe_loss}.

\begin{figure} 
    \centering
    \includegraphics[trim=7cm 1.5cm 6cm 1.5cm, clip, scale=0.35]{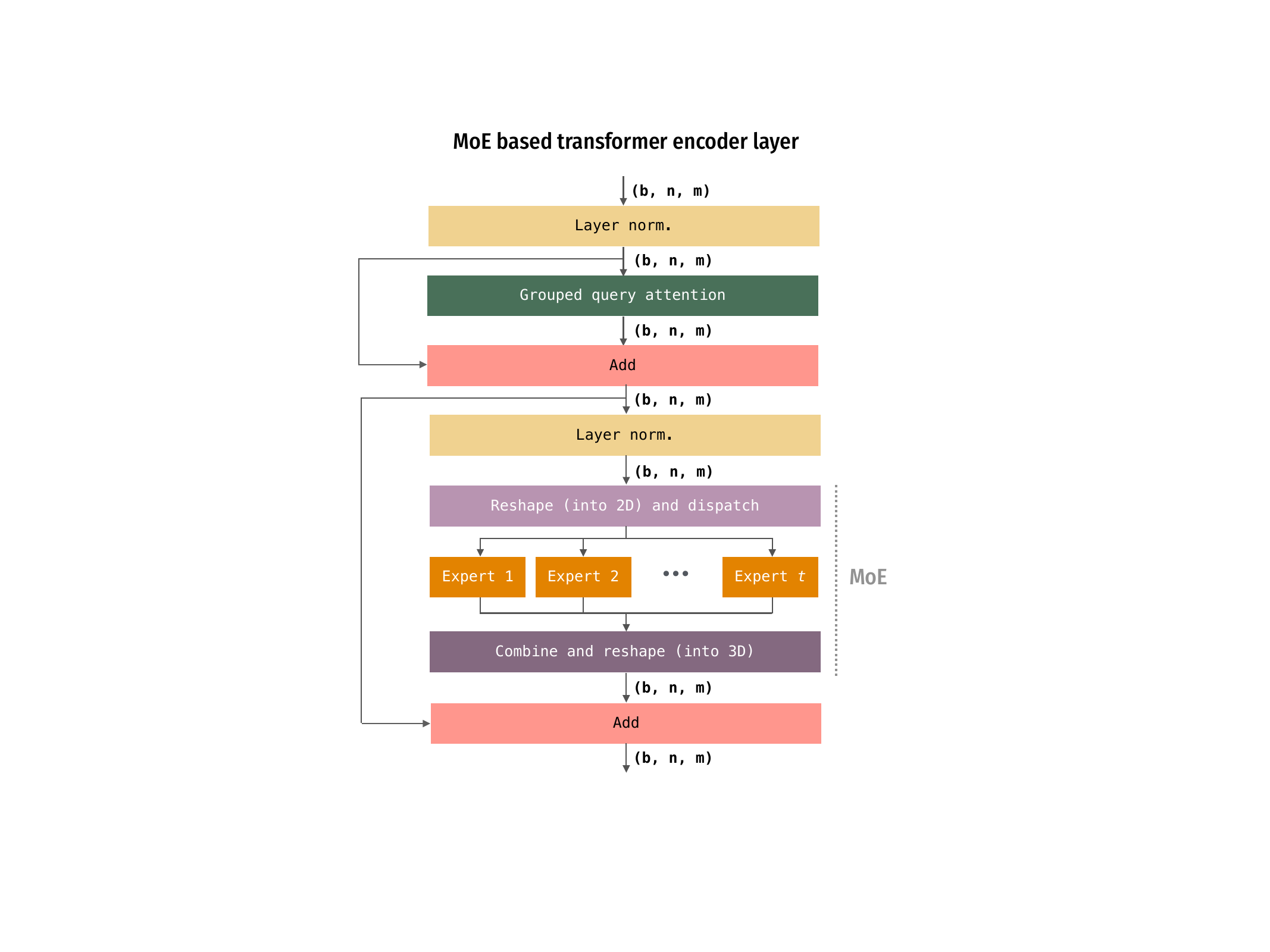}
    \caption{The architecture of \textbf{mmLiT}. $b$, $m$ and $p$ stand for batch size, embedding size and patch size, respectively. $n_E$ and $n_D$ are the number of embeddings/patches at the encoder and decoder respectively. $n_E + n_D = n$. $L_E$ is the total number of MoE transformer encoder layers at encoder; $L_D$ is the total number the layers at decoder. }
    \label{fig:mmLiT_layers}
\end{figure}

\section{Experimental Setup}
\label{sec:experiment}

We investigate the performance of \textbf{mLiT} and \textbf{mmLiT} at three different sizes: S, XS and XXS. Table \ref{tab:mLiT_config} shows the details of the encoder of each model, and Table \ref{tab:decoder_config} for decoder. For \textbf{mmLiT}, we perform self-supervised pre-training on the Cifar100 \cite{krizhevsky2009learning} training datasets and fine-tune on Cifar100, Cifar10\cite{krizhevsky2009learning}, Flowers102 \cite{flowers102} and Svhn \cite{svhn}. Additionally, we conduct supervised learning (\text{without any pre-training}) on the aforementioned four datasets.

The input image to the models has a size of 36 x 36, slightly larger than the original 32 x 32 dimensions of these datasets (except Flowers102, which is much larger and varied) . This results in each image being divided into 144 patches, given the patch size of 3 x 3. Similar to \cite{tan2024pretraining}, an auxiliary dummy patch is added to each image during both the pre-training and fine-tuning phases. This dummy patch, which contains zero values for all elements, is appended as the 145th patch for classification purposes. 

\begin{table}
\captionsetup{skip=5pt}
\centering
\caption{Configurations for various sizes of \textbf{mLiT}}
\begin{tabular}{lccc}
\toprule
\textbf{Configuration} & \textbf{S} & \textbf{XS} & \textbf{XXS} \\
\midrule
Embedding size & 144 & 128 & 108 \\
No. of layers ($L_E$) & 15 & 12 & 9 \\
Hidden size & 72-144 & 32-96 & 27-81 \\
No. of Attn. heads & 8 & 8 & 6 \\
No. of Attn. groups & 4 & 4 & 3 \\
No. of experts & 3-5 & 3-5 & 3-5 \\
k & 2 & 2 & 2 \\
No. of stages & 3 & 3 & 3 \\
No. of parameters & 2.36M & 1.21M & 0.66M \\
Dropout rate @ SwiGLU & 0.1 & 0.1 & 0.1 \\
\bottomrule
\end{tabular}
\label{tab:mLiT_config}
\end{table}

\begin{table}
\captionsetup{skip=5pt}
\centering
\caption{Configurations for decoder in \textbf{mmLiT}}
\begin{tabular}{lccc}
\toprule
\textbf{Configuration} & \textbf{S},\textbf{XS},\textbf{XXS} \\
\midrule
Embedding size & 108 \\
No. of layers ($L_D$) & 4 \\
Hidden size & 72 \\
No. of Attn. heads & 6 \\
No. of Attn. groups & 3 \\
No. of experts & 3 \\
k & 2 \\
No. of parameters & 0.34M \\
Dropout rate @ SwiGLU & 0.1 \\
\bottomrule
\end{tabular}
\label{tab:decoder_config}
\end{table}

All linear layers within the MoE based transformer encoder layers include bias. However, we exclude bias from other linear projection layers in both the encoder and decoder. For initializing weights and biases across all layer types, we rely on the default methods provided by Pytorch. The same applies to our approach to layer normalization \cite{ba2016layer}, where we use Pytorch's default setup.

\subsection{Pre-training}
\textbf{mmLiT}-S, \textbf{mmLiT}-XS, and \textbf{mmLiT}-XXS were pre-trained on Cifar100 for 4000, 6000 and 8000 epochs, respectively. We employed the AdamW optimizer \cite{loshchilov2018decoupled} with a weight decay set at 0.05. The initial 5\% of the total epochs were designated for warm-up \cite{goyal2018accurate}. We followed this with a cosine decay schedule \cite{loshchilov2017sgdr} for the learning rate and adhered to the linear learning rate scaling rule with a base learning rate of $3e-4$ \cite{goyal2018accurate, he2022masked}:

\begin{equation}
\mathrm{lr} = \mathrm{base\_lr} \times \mathrm{batch\_size} /256
\label{eq:lr}
\end{equation} See Table \ref{tab:pretrain_recipe} for more details.

\begin{table}
\captionsetup{skip=5pt}
\centering
\caption{Parameters and configuration for pre-training}
\begin{tabular}{ll}
\toprule
\textbf{Configuration} & \textbf{Value} \\
\midrule
Optimizer & AdamW \\
Weight decay & 0.05 \\
Base learning rate & $3 \times 10^{-4}$ \\
Learning rate schedule & Cosine decay \\
Warm-up epochs & 200 (S), 300 (XS), 400 (XXS) \\
Batch size & 840 (S), 1280 (XS, XXS) \\
Horizontal flipping & $p=0.5$ \\
Random resized cropping & $[0.6, 1]$ \\
Color normalization & $\mathrm{mean}=0.5, \mathrm{std}=0.5$\\
Discounting factor ($\alpha$) & 0.1 \\
MoE loss coefficient ($\beta$) & 0.5 \\
\bottomrule
\end{tabular}
\label{tab:pretrain_recipe}
\end{table}

\subsection{Fine-tuning}

For each pre-trained model, we conducted two sets of fine-tuning experiments. First, we fine-tuned the models pre-trained on Cifar100 for Cifar100 classification over 300 epochs. Second, we evaluated the transfer learning capabilities of the models by fine-tuning each one on Cifar100, Cifar10, Flowers102, and SVHN for 100 epochs. Table \ref{tab:finetune_set1} shows the configuration for the first set of fine-tuning, and Table \ref{tab:finetune_set2} shows the deviations in configurations with respect to Table \ref{tab:finetune_set1} for the second set of fine-tuning. 

Additionally, we evaluated the performance of various sizes of \textbf{mLiT} models (trained from scratch with only supervised learning). Table \ref{tab:supervised_config} shows the deviations in configurations for supervised learning on \textbf{mLiT} on the four datasets. We adhere to the linear learning rate scaling rule (Eq. \ref{eq:loss}) for all fine-tunings and supervised learning. All pre-trainings and fine-tunings were done using mixed precision (torch.float16).

\begin{table}
\captionsetup{skip=5pt}
\centering
\caption{Hyperparameters for the first set of fine-tuning}
\begin{tabular}{ll}
\toprule
\textbf{Hyperparameter} & \textbf{Value} \\
\midrule
Optimizer & AdamW \\
Weight decay & 0.05 \\
Base learning rate & $5 \times 10^{-3}$ \\
Learning rate schedule & Cosine decay \\
Layer-wise decay & 0.9 \\
Total epochs & 300 \\
Warm-up epochs & 20 \\
Batch size & 448 \\
Horizontal flipping & $p=0.5$ \\
Random resized cropping & $[0.8, 1]$ \\
AutoAugment &Policy for Cifar10 \\
Color normalization & $\mathrm{mean}=0.5, \mathrm{std}=0.5$\\
\bottomrule
\end{tabular}
\label{tab:finetune_set1}
\end{table}

\begin{table}
\captionsetup{skip=5pt}
\setlength{\tabcolsep}{3pt}
\centering
\caption{Hyperparameters deviations for the second set of fine-tuning. The learning rate is in $10^{-3}$.}
\begin{tabular}{lcccc}
\toprule
\textbf{Hyperparameter} & Flowers102  & Svhn & Cifar10 & Cifar100 \\
\midrule
Batch size & 16 & 256 & 256 & 128 \\
Learning rate & 10 & 2.5 & 2.5 & 5\\
Layer-wise decay & 0.9 & 0.9 & 0.8 & 0.9 \\
Warm-up epoch & 10 & 10 & 5 & 5\\
AutoAugment policy & Cifar10 & Svhn & Cifar10 & Cifar10 \\

\bottomrule
\end{tabular}
\label{tab:finetune_set2}
\end{table}

\begin{table}
\captionsetup{skip=5pt}
\setlength{\tabcolsep}{3pt}
\centering
\caption{Hyperparameters deviations for supervised learning on \textbf{mLiT} across four datasets. The learning rate is in $10^{-3}$.}
\begin{tabular}{lcccc}
\toprule
\textbf{Hyperparameter} & Flowers102  & Svhn & Cifar10 & Cifar100 \\
\midrule
Batch size & 16 & 256 & 256 & 128 \\
Learning rate & 10 & 1 & 1 & 2\\
Layer-wise decay & 1 & 1 & 1 & 1 \\
Warm-up epoch & 10 & 10 & 5 & 5\\
AutoAugment policy & Cifar10 & Svhn & Cifar10 & Cifar10 \\

\bottomrule
\end{tabular}
\label{tab:supervised_config}
\end{table}

\section{Result}
\label{sec:result}

\subsection{On Pre-Training}

Figure \ref{fig:mmLiT_loss} presents the pre-training losses for various sizes of \textbf{mmLiT}. The reductions in training losses exhibit approximately a log-linear relationship when plotted on logarithmic scales for both the x-axis (epochs) and y-axis (loss values). This indicates an exponential decrease in losses across epochs.

\textbf{mmLiT}-S exhibited the lowest loss, reaching 0.020404 at epoch 4000. \textbf{mmLiT}-XS and \textbf{mmLiT}-XXS achieved losses of 0.021051 and 0.021514 at epochs 6000 and 8000, respectively. The pre-training of \textbf{mmLiT}-S was notably stable, as evidenced by the minimal variability in its training loss. In contrast, the training losses for \textbf{mmLiT}-XS and \textbf{mmLiT}-XXS showed greater variability and more frequent, pronounced spikes, particularly noticeable after epoch 1000.

\begin{figure*} 
    \centering
    \includegraphics[trim=0cm 7.8cm 0cm 6cm, clip, scale=0.45]{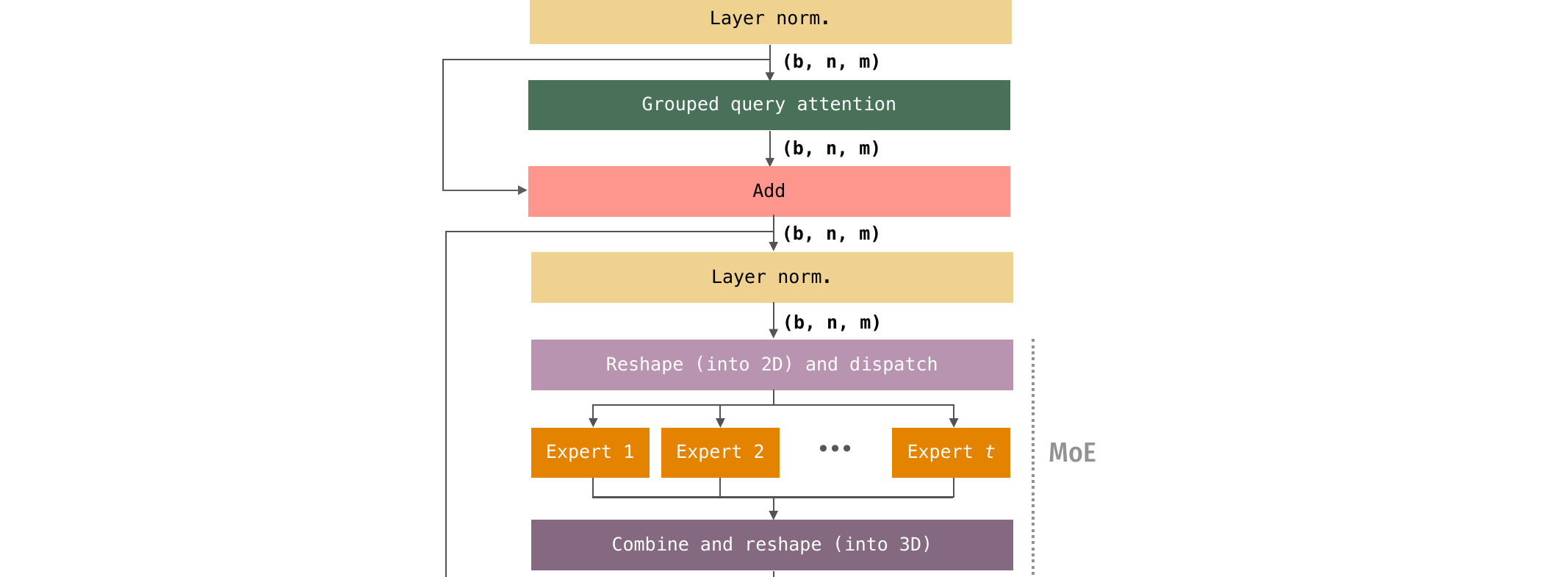}
    \caption{Pre-training losses for \textbf{mmLiT}-S,\textbf{mmLiT}-XS, and \textbf{mmLiT}-XXS. The pre-training of \textbf{mmLiT}-S was the most stable, with minimal variability. This can be observed by the thickness of the curves.}
    \label{fig:mmLiT_loss}
\end{figure*}

\subsection{On Fine-Tuning}
Table \ref{tab:result} compares the performances of models with similar sizes, all trained or fine-tuned over 300 epochs. \textbf{mmLiT}-S achieves an accuracy nearly 1\% higher than \textbf{Mae-ViT-C100}, despite having only two-thirds the parameters. \textbf{mmLiT}-XS, which is one-third the size of \textbf{Mae-ViT-C100}, exhibits a performance only 1.5\% lower. More interestingly, \textbf{mmLiT}-XXS, with just 18\% of the parameters of \textbf{Mae-ViT-C100}, remains competitive, trailing by only 3.3\% in accuracy. Furthermore, \textbf{mmLiT}-XXS slightly outperforms \textbf{ResNet56}, even though \textbf{ResNet56} has 18\% more parameters.

Table \ref{tab:result_transfer} illustrates the transfer learning capabilities of the models pre-trained only on Cifar100, with reference to the results reported from \cite{das-limiteddatavit-wacv2024}. It can be seen that \textbf{mmLiT} is competitive even at a scale of 0.67M parameters, where \textbf{ViT-T}+SSAT is almost 9 times larger, and the rests are at least 30 times larger than \textbf{mmLiT}-XXS. Furthermore, with the exception of \textbf{CVT-13}, which has convolutions in the architecture, \textbf{mLiT} always performs better than vanilla \textbf{ViT} and \textbf{Swin-T} even at the smallest scale on Cifar100, Cifar10 and Svhn. Flowers102 is a fine-grained classification dataset, so it is no surprise that \textbf{mLiT} is not competitive at an image size of 36 x 36.

\begin{table}
\caption{Comparison of Top-1 validation accuracy}
\setlength{\tabcolsep}{10pt}
\centering
\begin{tabular}{lrr}
\toprule
\textbf{Model}  & \textbf{Cifar100} & \textbf{\# Params}  \\
\midrule
\multicolumn{3}{l}{\textit{Convolutional Networks (Designed for CIFAR)}} \\
\midrule
\textbf{ResNet56} \cite{he2016deep}  & 74.81\% & 0.85 M  \\
\textbf{ResNet110}\cite{he2016deep}  & 76.63\% &  1.73 M  \\
\midrule
\multicolumn{3}{l}{\textit{Vision Transformers}\cite{hassani2021escaping}} \\
\midrule
\textbf{ViT-12/16}  & 57.97\% &  85.63 M  \\
\midrule
\textbf{ViT-Lite-7/16} & 52.87\% & 3.89 M  \\
\textbf{ViT-Lite-7/8} & 67.27\% & 3.74 M  \\
\textbf{ViT-Lite-7/4} & 73.94\% & 3.72 M  \\
\midrule
\multicolumn{3}{l}{\textit{Compact Vision Transformers}\cite{hassani2021escaping}} \\
\midrule
\textbf{CVT-7/8}  & 70.11\% & 3.74 M  \\
\textbf{CVT-7/4}  & 76.49\% & 3.72 M  \\
\midrule
\multicolumn{3}{l}{\textit{Compact Convolutional Transformers}\cite{hassani2021escaping}} \\
\midrule
\textbf{CCT-2/3 $\times$ 2}  & 66.93\% & 0.28 M  \\
\textbf{CCT-7/3 $\times$ 2}  & 77.72\% & 3.85 M  \\
\midrule
\multicolumn{3}{l}{\textit{MAE Vision Transformers} \cite{tan2024pretraining}} \\
\midrule
\textbf{Mae-ViT-C100}  & 78.27\% & 3.64 M  \\
\midrule
\multicolumn{3}{l}{\textit{\textbf{Current work on mmLiT}}} \\
\midrule
\textbf{mmLiT}-S & \textbf{79.15\%} & 2.38M \\
\textbf{mmLiT}-XS & 76.70\% & 1.22M \\
\textbf{mmLiT}-XXS & 74.95\% & 0.67M \\
\bottomrule
\end{tabular}
\label{tab:result}
\end{table}

\begin{table*}
\caption{Top-1 classification accuracy (in percentage) of various models on Cifar100, Cifar10, Svhn, and Flowers102. All models were trained for 100 epochs. For \textbf{mmLiT} and \textbf{mLiT}, all images were resized to 36 x 36, including Flowers-102. For the other models, the image size was set to 32 x 32 for Cifar100, Cifar10, and Svhn, and 224 x 224 for Flowers102.}
\centering
\begin{tabular}{l|S|ccc|c}
\toprule
\textbf{Model} & {\textbf{\# Param.} (M)} & \textbf{Cifar100} & \textbf{Cifar10} & \textbf{Svhn} & \textbf{Flowers102}  \\
\midrule
\multicolumn{6}{l}{\textit{Vanila variants of ViT} \cite{das-limiteddatavit-wacv2024}}  \\
\midrule
\textbf{ViT-T} \cite{touvron2021training} & 5.4 & 55.11 & 79.47 & 92.04 & 45.51 \\
\textbf{ViT-S} \cite{touvron2021training} & 21.4 & 54.08 & 79.93 & 94.45 & 56.17 \\
\textbf{CVT-13} \cite{haipingcvt} & 20. & 73.50 & 89.02 & 91.47 & 54.29 \\
\textbf{Swin-T} \cite{swintransformer} & 29. & 53.28 & 59.47 & 71.60 & 34.51 \\
\midrule
\textbf{ResNet-50} \cite{he2016deep, das-limiteddatavit-wacv2024} & 25.6 & 72.80 & 91.78 & 96.45 & 46.92 \\
\midrule
\multicolumn{6}{l}{\textit{\textbf{mLiT} (no pre-training)}} \\
\midrule
\textbf{mLiT}-S & 2.38 & 60.58 & 84.03 & 96.01 & 34.41 \\
\textbf{mLiT}-XS & 1.22 & 57.93  & 82.74 & 95.43 & 35.41 \\
\textbf{mLiT}-XXS & 0.67 & 56.98  & 80.78 & 94.97 & 35.34 \\
\midrule
\multicolumn{6}{l}{\textit{Variants of ViT augmented by Masked Autoencoders} \cite{das-limiteddatavit-wacv2024}}  \\
\midrule
\textbf{ViT-T}+SSAT & 5.8 & 69.64 & 91.65 & 97.52 & 57.20 \\
\textbf{ViT-S}+SSAT & 21.8 & 73.37 & 94.05 & \textbf{97.87} & 61.15 \\
\textbf{CVT-13}+SSAT & 20.3 & 75.16 & \textbf{95.93} & 97.00 &\textbf{68.82} \\
\textbf{Swin-T}+SSAT & 29.3 & 60.68 & 83.12 & 85.83 & 54.72 \\
\midrule
\multicolumn{6}{l}{\textit{\textbf{mmLiT} (pre-trained on} Cifar100 \textit{)}} \\
\midrule
\textbf{mmLiT}-S & 2.38 & \textbf{78.18}  & 94.62 & 97.39 & 59.59 \\
\textbf{mmLiT}-XS & 1.22 & 75.91  & 94.01 & 97.26 & 57.46 \\
\textbf{mmLiT}-XXS & 0.67 & 73.42  & 92.21 & 97.32 & 48.59 \\
\bottomrule
\end{tabular}
\label{tab:result_transfer}
\end{table*}

\section{Discussion}
The results from Table \ref{tab:result_transfer} show that ViT can learn better with streamlined MoE at a much smaller scale. In our current proposal, $\mathbf{V}$ and $\mathbf{W_2}$ are shared across an MoE layer. The main consideration behind this arrangement is the non-linear $\mathrm{silu}$ applied on $\mathbf{xW}$. Nevertheless, we have also explored two other possible sharing arrangements on \textbf{mmLiT}-S (see Table \ref{tab:items_shared}), revealing that retaining $\mathbf{W}$ gives a slight advantage. 

\begin{table}
\captionsetup{skip=5pt}
\centering
\caption{Investigation into various ways of sharing linear transformations in SwiGLU across MoE layer}
\begin{tabular}{lc}
\toprule
\textbf{Items shared} & \textbf{Acc. @ Cifar100} \\
\midrule
$\mathbf{V}$, $\mathbf{W_2}$ & 79.15 \\
$\mathbf{V}$, $\mathbf{W}$   & 78.34 \\
$\mathbf{W}$, $\mathbf{W_2}$   & 78.87 \\
\bottomrule
\end{tabular}
\label{tab:items_shared}
\end{table}

In our design, we have progressively reduced the hidden size and increased the number of experts at several stages. This arrangement is inspired by convolutional neural networks \cite{krizhevsky2012imagenet}, where the size of the feature maps is reduced successively across layers, and the number of feature maps increases as the network gets deeper. We found that with the arrangement of MoE, a reduction in hidden size leads to little deterioration in performance if the reduction is not significant at the early layers and not more than 75\% in the last layers. Further experiments are required in future to confirm this.

When we performed fine-tuning on the four datasets, we started by applying the fine-tuning setup we used for Cifar100 to the other three datasets. But overfitting at an early stage became a consistent issue, and because of that, we had to modify the setup to accommodate each dataset. We encountered similar problem when we tried to perform supervised learning with \textbf{mLiT} from scratch. Therefore, we reported setups in Table \ref{tab:finetune_set1} and Table \ref{tab:finetune_set2} for clarity. It is worth nothing that we did not perform a thorough search on the configurations; we simply adjusted the hyperparameters to avoid overfitting during training. Furthermore, unlike in \cite{das-limiteddatavit-wacv2024}, we did not use advanced augmentation techniques in all our fine-tunings.

During pre-training of \textbf{mmLiT}, we used different amount of epochs for the different sizes of models. From Figure \ref{fig:mmLiT_loss}, it is clear that more epochs should have been allocated for \textbf{mmLiT}-XS and \textbf{mmLiT}-XXS from the perspective of loss. However one of the downsides of masked autoencoders is the unclear relationship between the loss of a model and its subsequent performance in downstream tasks. In our experience, we have seen cases where further pre-training for a few hundreds epochs at a later stage leads to similar or poorer performance in classification. This might be a specific issue when the dataset is small. Despite the above, we believe that with such datasets, further pre-training of thousands of epochs should lead to a better model. 

In this study, we found that \textbf{mLiT} pre-trained on only 50,000 images can serve as a sort of foundation model \cite{Bommasani2021FoundationModels} even at the smallest size. On both Cifar10 and Flowers102, models pre-trained on Cifar100 improve by at least 10\%. On Cifar10, \textbf{mmLiT}-XXS can reach 90\% accuracy in less 40 epochs. On Svhn, the improvements are modest. This is partly because \textbf{mLiT} is already performing well without pre-training. It is also possibly due to the lack of similar images in Cifar100. We believe a model pre-trained on a slightly larger and a more diverse dataset can perform competitively on various simpler tasks at these tiny scales.

In literature, it is common to integrate convolutional layers \cite{graham2021levit, hassani2021escaping} or employ a convolutional 'teacher' \cite{touvron2021training} for imparting inductive bias. It seems the use of streamlined MoE can help alleviate the lack of inductive bias. However, as demonstrated in \cite{tan2024pretraining, das-limiteddatavit-wacv2024}, with a masked auto-encoder setup, the problem of inductive bias can be overcome.

\section{Conclusion}
In this paper, we demonstrated the potential of streamlined MoE architecture to create very lightweight vision transformers. By sharing parameters across MoE layers and adopting depth-wise scaling, we achieved a balance between model complexity and performance. Furthermore, our findings suggest that pre-training on a slightly larger and more diverse dataset could enhance the model's versatility and efficacy across various tasks. The streamlined MoE approach appears promising in mitigating the lack of inductive bias, particularly when used in conjunction with masked autoencoder setups.

\bibliographystyle{unsrt}  
\bibliography{references}

\end{sloppy}
\end{document}